# Population size predicts lexical diversity, but so does the mean sea level – why it is important to correctly account for the structure of temporal data.

**Alexander Koplenig**[1*], **Carolin Müller-Spitzer**[1]

**1** Department of Lexical Studies, Institute for the German language (IDS), Mannheim, Germany.

* Corresponding author

E-mail: Koplenig@ids-mannheim.de

## Abstract

In order to demonstrate why it is important to correctly account for the (serial dependent) structure of temporal data, we document an apparently spectacular relationship between population size and lexical diversity: for five out of seven investigated languages, there is a strong relationship between population size and lexical diversity of the primary language in this country. We show that this relationship is the result of a misspecified model that does not consider the temporal aspect of the data by presenting a similar but nonsensical relationship between the global annual mean sea level and lexical diversity. Given the fact that in the recent past, several studies were published that present surprising links between different economic, cultural, political and (socio-)demographical variables on the one hand and cultural or linguistic characteristics on the other hand, but seem to suffer from exactly this problem, we explain the cause of the misspecification and show that it has profound consequences. We demonstrate how simple transformation of the time series can often solve problems of this type and argue that the evaluation of the plausibility of a relationship is important in this context. We hope that our paper will help both researchers and reviewers to understand why it is important to use special models for the analysis of data with a natural temporal ordering.

## Introduction

In principle, we could start our paper like this: As complex adaptive systems [1], languages are constantly changing on all fundamental levels [2]. In this context, several studies have indicated that population size is an important factor that influences the (rate of) language change [3–5]. Based on a very large sample of the written language record [6], we present quantitative evidence that clearly supports the claim that population size strongly influences the lexical diversity of eight languages. The problem is: an almost identical pattern emerges if we correlate the lexical diversity in a given year with the global mean sea level [7] instead of population size. This strongly raises the suspicion that there might be something wrong with our analysis.. The description of this problem and its consequences are at the very heart of our study: we show that whenever two variables evolve through time, those variables will almost always look highly correlated even if they are not related in any substantial sense. While we want to point out upfront that the problem is fairly common knowledge in econometric time-series analysis [8–11], we nevertheless felt the need to outline it once again, in an instructive and non-technical manner, in this paper, given the fact that in the recent past, several studies were published– some in journals with a good reputation – that seem to suffer from exactly this problem [12–18]. Following [19], we consider the fact that those studies do not correctly account for the temporal structure of the data problematic, because the studies sometimes present surprising or even spectacular relationships, especially between different economic, cultural, political and (socio-)demographical variables on the one hand and cultural or linguistic characteristics (like the one described above) on the other hand. As a result, the studies also receive widespread media attention and can, in turn, affect public opinion and decision making. We therefore hope that our paper can help both researchers and reviewers detect the problem and ultimately avoid it.

The remainder of this paper is organized in the following way: First, we document an apparently spectacular relationship between population size and lexical diversity. We then try to show that this relationship is the result of a misspecified model that does not correctly account for the (serial dependent) structure of temporal data of the data by presenting a similar but nonsensical relationship between the global annual mean sea level and lexical diversity. On this basis, we explain the cause of the misspecification, show that it has profound consequences and demonstrate how simple data transformation can often solve the problem, before arguing that checking for plausibility is important in this context. This paper ends with some concluding remarks.

## Results and Discussion

### The problem of spurious correlations

We 'investigated' the correlation between population size and lexical diversity for American English, British English, Chinese (simplified), French, German, Italian, Russian and Spanish on the basis of data on population size and the type-token ratio based on the Google Books dataset (see Materials and Methods for more details). At first glance, Fig 1 seems to document a spectacular and fascinating relationship: There is a strong and statistically significant correlation (at $p < .05$ or better) between the level of the population size and the level of the type-token ratio as a measure of lexical diversity for all investigated languages, except Chinese. On this basis, we could argue for a general relationship. For example, we assume that with increasing populations, the number of language speakers naturally also increases. We could then continue and elaborate our argument by assuming that a larger number of speakers is most likely associated with a greater degree of variance in demographic background and sociocultural environments [20], and that this greater diversity leads to an increase of the type-token ratio as a measure of lexical diversity. The fact that there is virtually no correlation for Chinese and – compared to the other languages – only a small correlation for the Russian data could then be incorporated in our "theory" by referring to the political background in those two countries, e.g. that in socialist countries, increasing population sizes do not affect the lexical diversity, because socialist coercion policies suppresses linguistic development by defining one common "linguistic" or "cultural" standard or something like that. This is exactly what [19] alerted to, since this result could be used in the media or even by politicians to criticize socialism.

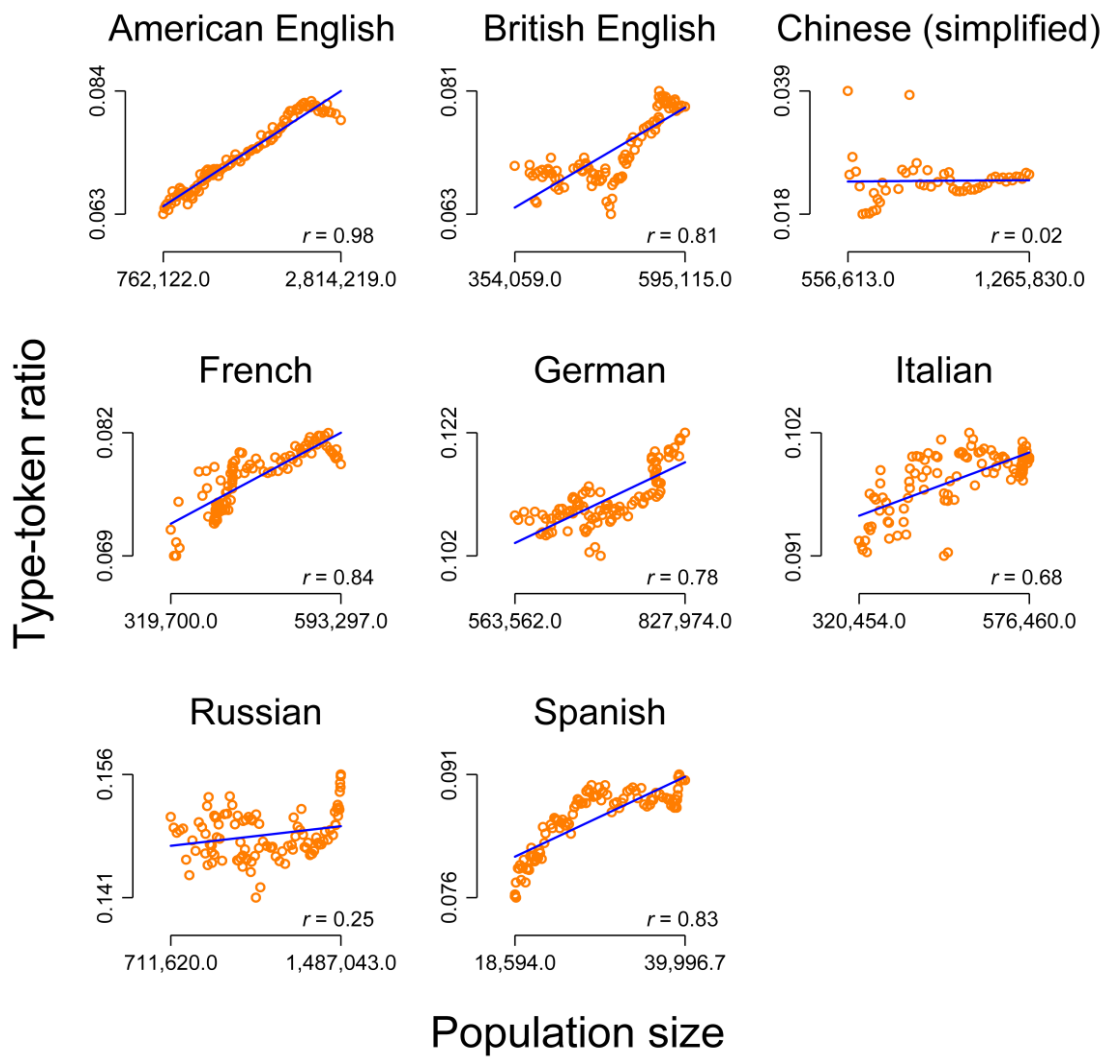

**Fig 1: Correlation between the level of the population size and the level of the type-token ratio for eight languages (including two varieties of English)**. Orange circles: raw data. Blue line: linear prediction of the type-token ratio on the population size. Notes on the bottom right: Pearson correlation (all *p*s < .05 or better, except for Chinese where *p* = .89).

Fig 2 demonstrates why the analysis presented in Fig 1 is severely flawed: Compared to the cross-linguistic relationship between population size and lexical diversity, which it would be possible to argue for, the relationship between the global mean sea level and the lexical diversity is highly similar. In this context, all correlations except for Chinese and Russian are strong and highly significant (at *p* < .001 or better). However, arguing for any potential relationship seems to be out of the question in this context.

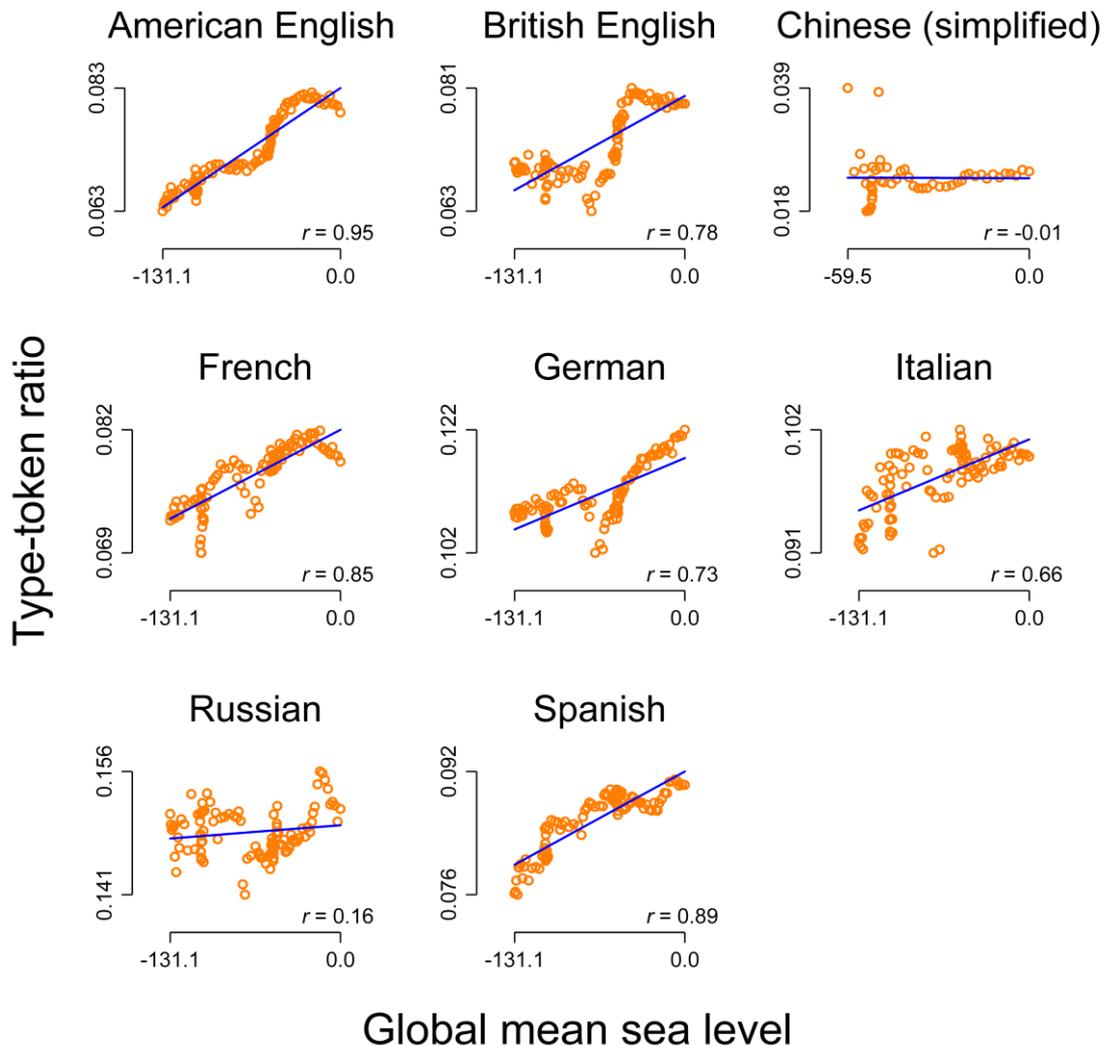

**Fig 2: Correlation between the level of the annual global mean seal level and the level of the type-token ratio for eight languages (including two varieties of English).** Orange circles: raw data. Blue line: linear prediction of the type-token ratio on the population size. Notes on the bottom right: Pearson correlation (all *p*s < .001 except for Chinese were *p* = 0.95 and Russian where *p* = .10)

Fig 3 "explains" these apparent relationships: all type-token ratio time-series, again except for Chinese and Russian, clearly exhibit an upward trend. The series are said to have a unit root or to be non-stationary [11]. The same is true for both the population

sizes and the global mean sea level, which also increased throughout the 20$^{th}$ century. For the analysis of temporal data, this has important ramifications because the following statement is true *per definition*: values that are later in time will be above the average of the mean value of the series, while values that are earlier in time will be below average. Since the Pearson product-moment correlation measures whether values of one series that are above/below average tend to co-occur with values of another series that are above/below average, by mathematical necessity, the correlation coefficient for two trending time-series will then be high when in fact they are not related in any substantial sense [8]. An augmented Dickey-Fuller test (a formal test for a unit root) with a lag length of 1 reveals that all time series that include the population sizes and the global mean sea level are non-stationary (all *p*s > .05), except for the Chinese and the Russian type-token ration series where *p* < .05.

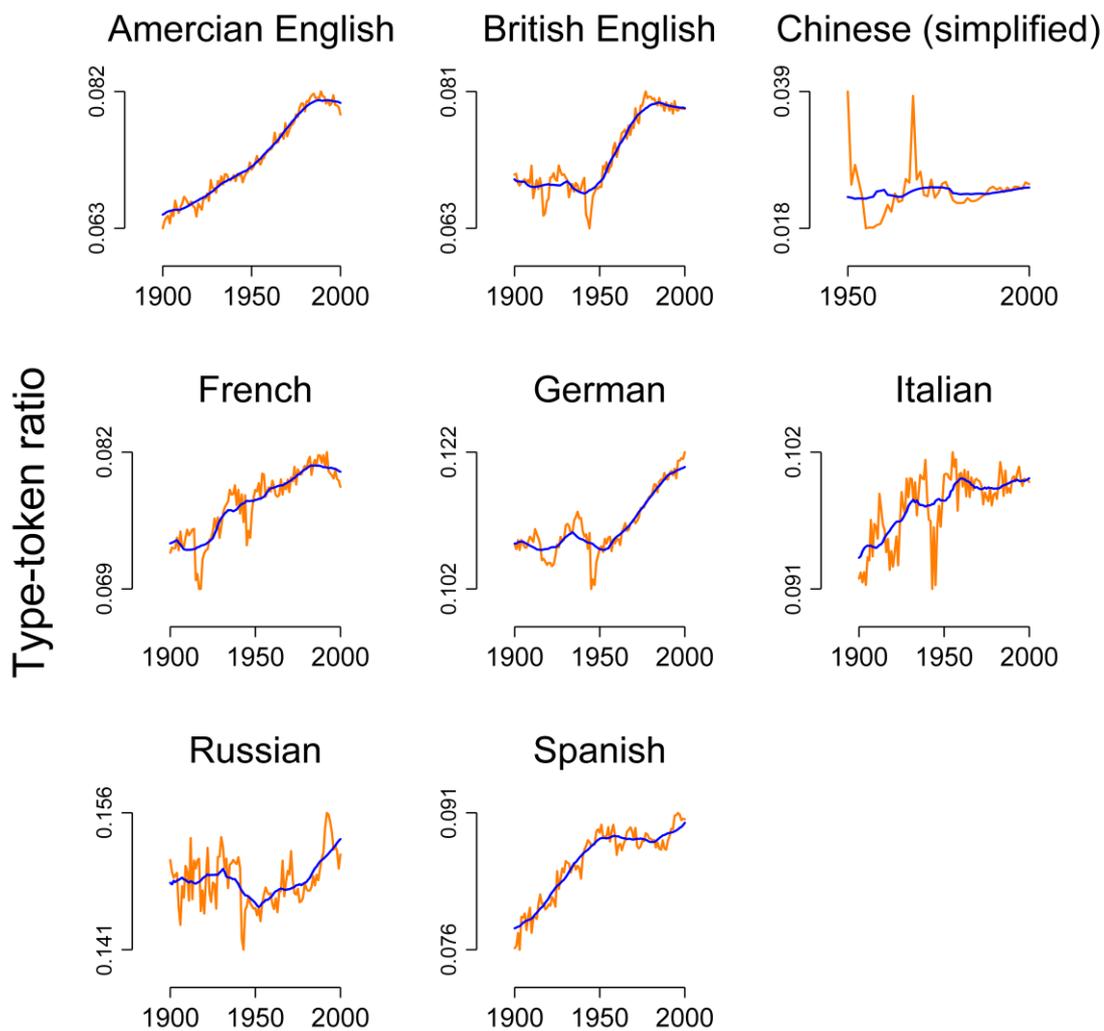

**Fig 3: Type-token ratio as a function of time for eight languages (including two varieties of English).** Orange lines: raw data. Blue lines: simple weighted moving average with an 11-year window centered on the current value.

## Changes instead of levels

One common approach to avoid spurious correlations is to transform the series prior to the analysis, for example by detrending the series (estimating the trend and subtracting it from the actual series). Another more general solution that often results in stationary series, that is a series in which the mean and the variance of the investigated series do not change as a function of time, is to correlate period-to-period changes instead of the actual levels of the two series [11]. It is worth emphasizing that this also re-formulates the research question as it excludes the upward trends of the correlated series [21]. It determines if period-to-period changes that are above or below the average of the first series correspond mainly to changes that are above or below the average of the second series. Therefore, [8] suggest as a rule of thumb to generally model data on a combination of both levels and changes. In our case, correlating year-to-year changes (or decade-to-decade changes) seems to be even better suited to answer our "research question": if the population increases from last year to this year, then – on average –lexical diversity should also increase from last year to this year, if both series are related. If we correlate year-to-year changes all correlations between population size / global mean sea level and lexical diversity become virtually nonexistent ($r_{max}$ = .10) and insignificant at all common levels of significance ($p_{min}$ = .31).

## The problem of temporal autocorrelation

To demonstrate why it is problematic to correlate two trending time-series, we have simulated 10,000 random walks with drift (cf. Materials and Methods). Each resulting time-series has an average upward trend, but otherwise behaves in a completely random manner. This means that the random walks serve as a proxy for time series with a general upward trend. All series are then correlated with the annual global mean sea level. Fig 4 shows that 7,519 of the 10,000 random walks correlate moderately ($r$ > .30) and more than 50% (5,432) even correlate strongly ($r$ > .75) with the global mean sea level, if we calculate the correlation based on the level of the series ($r_{mean}$ = .52). This result is, of course, far from what we should actually expect for the distribution of correlation coefficients where one variable is a random quantity *per definition*: only a few series should – by chance – substantially correlate with the global mean sea level, while most correlation coefficients should be close to zero. If we instead correlate year-to-year changes of the two series, the maximum correlation coefficients amounts to $r = .37$ and – as expected – a distribution that closely resembles a normal distribution (blue line in Fig 4) with a mean close to zero ($r_{mean}= .00$).

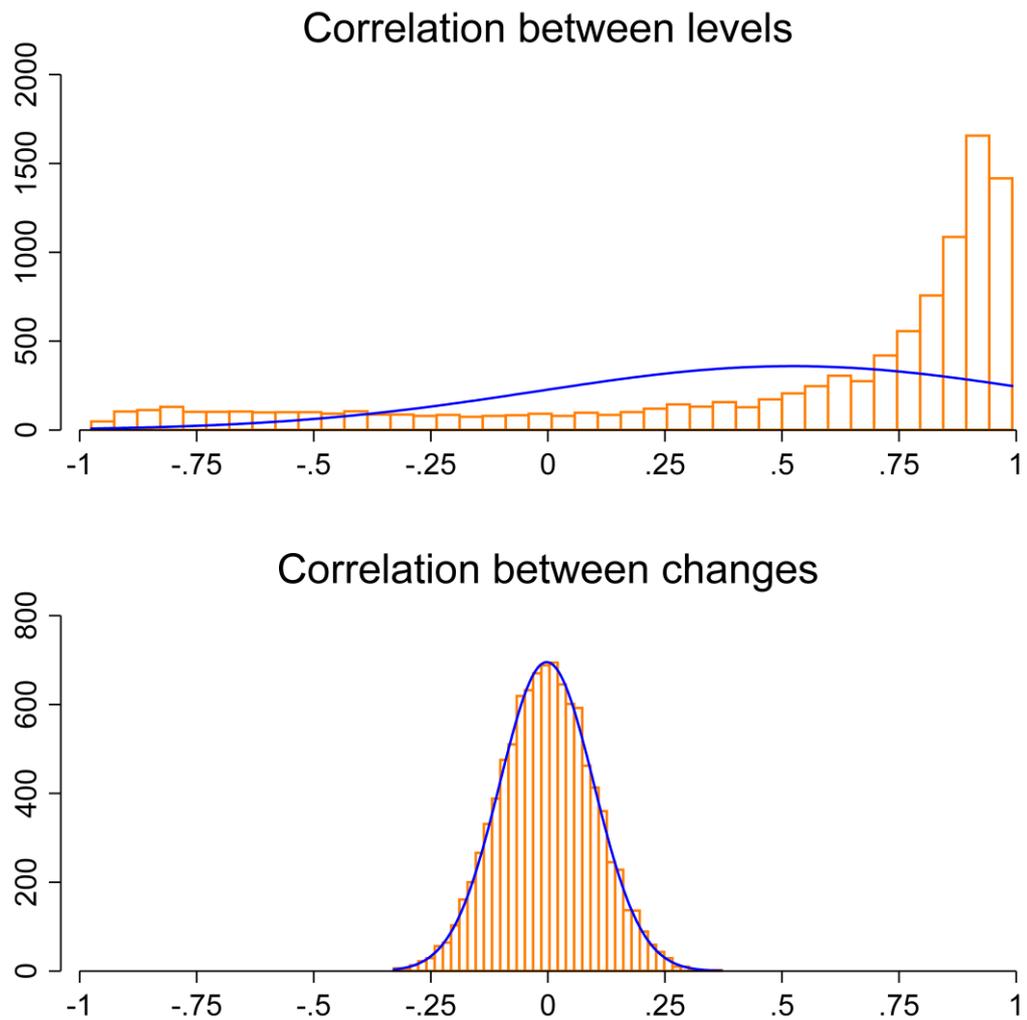

**Fig 4: Correlations between the annual global mean seal level and 10,000 simulated random walks with drift.** Top: Histogram of the correlations between levels. Bottom: Histogram of the correlation between year-to-year changes. The height of the bars in both histograms represents the number of cases in the category. Blue lines: scaled normal density.

This leaves little room for debate: Whenever two variables evolve through time, those variables will almost always look highly correlated even if they are not related in any substantial sense. The reason why standard statistical models fail when it comes to the analysis of time-series has to do with the fact that there is basically no such thing as a univariate time-series: analyzing univariate time series is always "the analysis of the bivariate relationship between the variable of interest and time." [11, p. 92]. In the Materials and Methods section, we demonstrate why temporal autocorrelation is problematic from a statistical point of view, if regular models for cross-sectional data (that assume independence between individual observations) are used.

## Plausibility

To demonstrate why it is also important – especially for a spectacular and unexpected result – to remain skeptical and to carefully check plausibility, let us briefly give an example what our initial "analysis" of the relationship between lexical diversity and population size would actually imply: if we regress the level of the lexical diversity in the Spanish Google Books data on the population size of Spain, we obtain a coefficient of determination of $r^2 = .69$. This means that almost 70% of the variance of the lexical diversity variable is "explained" by the population size (for the American English data it would be even more than 95% of the variance). This model would also imply that every 10 new inhabitants of Spain are equal to 4.56 additional word types (per 1 million word tokens) in the Spanish Google Books data (that also includes books written and published in Latin America). We believe that this would be an extraordinary result. In fact, this result would be so extraordinary that it seems wise to first ask: is this result plausible? Can we come up with any good theory regarding this relationship?

A few words on the Google Books data are in order here, as they are the basis of all but one [16] study quoted above. Here, we want to echo [22, p. 1203]: just because a dataset is big, does not mean that "one can ignore foundational issues of measurement and construct validity and reliability and dependencies among data." However, this seems to be precisely the case regarding the Google Books Ngram data. After all, for *n*-grams where *n* is ranges from one (single words) to five (five word units), the data consist of only year-wise aggregated overall frequencies of occurrence and the number of books each n-gram appears in. Since n-grams do not occur independently across distinct books, this aggregation of individual book frequencies means that we cannot account for the distributions of n-grams which can have profound consequences for the analysis of textual data [23,24]. In addition, it is a largely overlooked fact that the Google Books Ngram data only includes the counts for n-grams that occur at least 40 times across the entire corpus. At least from a (corpus) linguistic point of view, this certainly matters since most n-grams are very infrequent. So in terms of what we know about word frequency distributions [25], this procedure eliminates approximately 95% of all different 1-gram types; for n-grams where n>1 this figure is even higher [26]. To the best of our knowledge, the question of whether this arbitrary data truncation does not impose a systematic bias on the data is something that remains to be demonstrated empirically, given the fact that corpus size for each year strongly increases as a function of time. At the same time, this means that we would have to further extend our analysis illustrated above: nearly every second new inhabitant of Spain is "responsible" for one new word type *that occurs more than 40 times* in the Spanish Google Books data.

To check the plausibility of this result, we would have to face the fact that we still do not have any reliable information about the books included in the corpora. According to the FAQs of the Culturomics project behind the GB data [27] the team has "not received permission […] yet" to release the full 5.2 million book bibliography, containing infor-mation about each book included in the corpus for each language. This statement has not changed in the last few years and we are rather pessimistic that it will change anytime soon; but we would love to be proven wrong. This, in turn, means that we currently have no way of finding out whether the different diachronic book samples really represent similar things at different moments in time and, as the Culturomics team themselves pointed out [28, p. 13], it is very likely that the types of books that are published are changing as a function of time. The lack of metadata can have important ramifications for any interpretation of potential results based on the Google Books Ngram data [29,30]. So at the moment, all we can be sure about, again according to the aforementioned FAQs, is that "the vast majority of the books from 1800-2000 come from Google's library partners, and so the composition of the corpus reflects the kind of books that libraries tend to acquire."

Returning to our research question – the correlation between population size and lexical diversity – population growth is affected by the birth of children and the influx of immigrants. Babies do not write books, and only a few immigrants publish books which are acquired by libraries shortly after immigration. So, the strong relationship between lexical diversity and population size would indicate that nearly every second new inhabitant (babies and immigrants alike) is "responsible" for one new word type *that occurs more than 40 times* in the Spanish Google Books data. This would really be an extraordinary relationship.

To drive home this point, if we regress the level of lexical diversity in the German Google Books data to the population size of China, we obtain a very strong correlation of $r = .89$ that is significant at all standard levels. Our model predicts that with every 1,000 new inhabitants of China, we will find roughly 15 additional word types in the German Google Books data. While it is certainly possible to "rationalize nearly every-thing" [9, p. 2], we just do not think that this result makes any sense – which mechanism could generate a relationship like this? If we use year-to-year changes instead of the actual levels, we obtain an insignificant correlation of 0.10 ($p = 0.34$). This implies that knowing the Chinese population size does not help in predicting the lexical diversity in the Google Books data, a result which we believe fits reality more closely.

From a statistical point of view, this demonstrates why it can be a good idea to model a potential relationship between two trending time series with changes instead of levels. This is also important from a methodological point of view: just because two series are trending, does not necessarily imply any substantial relationship [31]. Therefore we strongly advise against using the fact that two series are evolving in a predicted way as evidence in order to substantiate a specific theoretical claim.

The general question concerning the Google Books data itself, whether the acquisition strategy of major libraries really can serve as an (temporarily) unbiased proxy for the evolution of subjective or even latent cognitive traits, is an open research question. Again, we are rather skeptical. For example, a change in the acquisition strategy of one major library is not necessarily motivated by one of the factors we might be interested in; nevertheless in aggregation of the frequency counts of different n-grams, it might look like one. Once again, we want to refer to [22, p. 1203] regarding such naïve mapping:

"All empirical research stands on a foundation of measurement. Is the instrumentation actually capturing the theoretical construct of interest? Is measurement stable and comparable across cases and over time? Are measurement errors systematic?"

The outlined problems all have to do with the fact that – in making the data freely available (which is a fantastic thing) – Google wanted to avoid breaking any copyright laws, and it goes without saying that legal restrictions also have to be taken seriously in this case. However, while we are – as many other empirically-minded researchers – fascinated by the possibilities that the analysis of "big data" offers, we believe that the seemingly prevailing view that the size of the (Google Books Ngram) data will stand in for fundamental methodological problems, is not justified.

## Concluding remarks

All recently published studies that we mentioned in the introduction do not explicitly model the underlying temporal structure of the data [12–18]. This certainly has to do with the fact that time series analysis is a relatively young statistical discipline [11, p. xxi-xxii]. To improve the reliability of research [32], we hope that this paper will help both researchers and reviewers to understand why it is important to use special models for the analysis of such data. Standard statistical models that work for cross-sectional data run the risk of incorrect statistical inference in time-series analysis, where (potentially strong) effects are meaningless and therefore can potentially lead to wrong conclusions.

While our analysis indicates that type-token ratios do not dependent on population sizes, this does not imply, of course, that the increase of the type-token ratios over time is not interesting in itself as Harald Baayen (personal communication) points out, because this increase could reflect the fact that onomasiological needs increase with the complexity of modern societies [33]. Or put differently, new ideas and new technologies need new designations in order to efficiently communicate about related concepts. Thus, under the assumption that cultural adaption is cumulative [34,35], a rapid increase of technological innovations could result in an increase of the type-token ratio, independently of the population size. This is certainly an interesting avenue for future research.

# Acknowledgments

We would like to thank Sascha Wolfer for valuable comments on earlier drafts of this article and Sarah Signer for proofreading. Also, we are grateful to Harald Baayen for insightful comments and additional inputs on the interpretation of our analyses as mentioned in the text.

# Materials and Methods

S1 contains the population data, compiled from [36]. The time-series of the global mean sea level was presented in [7] and is available at [37]. The type-token ratio, a common way to measure lexical diversity, is based on the Google Books datasets that were made available by [6] at [27]. For the present study, the 1-gram datasets of Version 2 (July 2012) of the following languages were used: American English, British English, Chinese (simplified), French, German, Italian, Russian and Spanish. The type-token ratio for each year and each language is calculated by dividing the number of unique strings by the total number of strings. Higher type-token ratios are indicative of higher lexical diversity. Since this measure is known to be heavily text-length dependent [38] and given the fact that the corpus size based on the Google Books data strongly increases as a function of time, calculating the type-token ratio based on the actual corpus sizes would systematically bias the results. To solve this problem, random samples of 1,000,000 tokens were drawn from the data as described in [39]. The analysis is restricted to the 20$^{th}$ century, except for Chinese, which is restricted to the time span 1950-2000 since the size of the Google Books base corpora is not sufficient (< 1,000,000 tokens) for earlier periods.

Additionally, we simulated $i = 10{,}000$ random walks with drift [40,11] that are defined as:

$$x_{t,i} = d_i - x_{t-1,i} + e_{t,i}$$

where $x_{t,i}$ is the value of the $i$th random walk at time point $t$; the constant drift term $d_i$ is randomly drawn from a uniformly distributed interval [0.02,0.2) and $e_{t,i}$ is white noise, normally distributed over the interval [0,1).

For each resulting series this means that the current value of the series depends on its previous value plus a positive drift term and a white noise error term. At each point in time, the series takes one random step away from the last position, but as result of the drift term, the series will have an upward trend in the long-run.

All analyses were carried out using Stata/MP2 14.0 for Windows (64-bit version). To ensure maximal replicability, S2 contains a Stata script ('do-file') that automatically downloads the data and reproduces all results presented in this article, while S3 is a delimited text file (comma-separated) of the final dataset that can be used to replicate our findings with another software package.

## Temporal autocorrelation

From a statistical point of view, temporal autocorrelation is problematic because it biases our estimators. If, for example, we fit a simple time-series regression that can be written as:

$$y_t = \beta_0 + \beta_1 x_{1t} + \varepsilon_t$$

where $y_t$ represents the level of our outcome variable in $t$ and $x_{1t}$ is the level of predictor variable, $\beta_0$ is the regression constant and $\beta_1$ is the regression coefficient, $\varepsilon_t$ is the error term. OLS analysis assumes that there is no autocorrelation between the residuals

$(Cov(\varepsilon_s, \varepsilon_t) = 0$ for all $s \neq t$). In this context, first-order autocorrelation $\varepsilon_t$ can be written as:

$$\varepsilon_t = \rho\varepsilon_{t-1} + \eta_t$$

where $\rho$ is the autoregressive parameter and $\eta_t$ is a white-noise process. In the presence of first-order autocorrelation, the OLS estimators are biased and lead to incorrect statistical inferences [41]. To see, why this is also the problem of our simulation, Fig 5 shows the correlation between current and lagged residuals of an OLS regression for each of the simulated random walks with drift on the annual global mean seal level. While the regression residuals of levels are strongly autocorrelated ($r_{mean}$= .91), we obtain a normal distribution with a mean close to zero ($r_{mean}$= -.02) for the regression residuals of year-to-year changes.

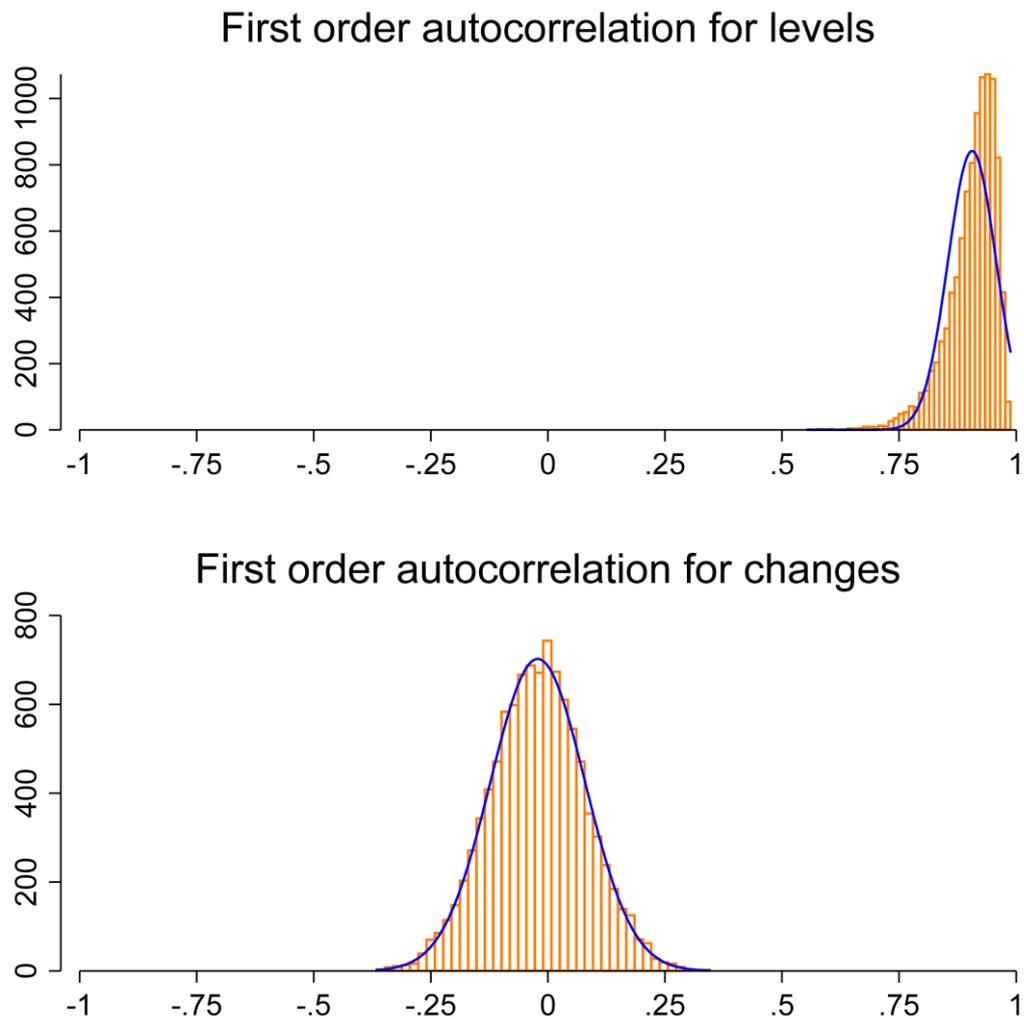

**Fig 5: Correlations of current and lagged residuals of a regression of the 10,000 simulated random walks with drift on the annual global mean seal level.** Top: Histogram for levels. Bottom: Histogram for year-to-year changes. The height of the bars in both histograms represents the number of cases in the category. Blue lines: scaled normal density.

# Supporting Information

S1 File. Population size data, compiled from [36].

S2 File. Stata do-file that automatically downloads the data and reproduces all results presented in the article.

S3 File. Delimited text file (comma-separated) of the final dataset that can be used to replicate our findings.